\documentclass{article}
\usepackage{ijcai24}

\usepackage{times}
\usepackage{soul}
\usepackage{url}
\usepackage[hidelinks]{hyperref}
\usepackage[utf8]{inputenc}
\usepackage[small]{caption}
\usepackage{graphicx}
\usepackage{amsmath}
\usepackage{amsthm}
\usepackage{booktabs}
\usepackage{algorithm}
\usepackage{algorithmic}
\usepackage[switch]{lineno}
\usepackage{twemojis}
\usepackage{tabularx}
\usepackage{verbatim}

\usepackage{lipsum} 
\usepackage{multicol} 

\usepackage[many]{tcolorbox}    	
\usepackage{setspace}               
\usepackage{multicol}               

\definecolor{main}{HTML}{555555}    
\definecolor{sub}{HTML}{cde4ff}     

\tcbset{
    sharp corners,
    colback = white,
    before skip = 0.2cm,    
    after skip = 0.2cm      
}                           

\newtcolorbox{boxB}{
    fontupper = \color{main}, 
    boxrule = 1.0pt,
    colframe = main,
    rounded corners,
    arc = 5pt   
}

\title{Efficient and Accurate Memorable Conversation Model using DPO based on sLLM}

\author{
Youngkyung Seo\and
Yoonseok Heo\and
Jun-Seok Koh\And
Du-Seong Chang\footnote{Corresponding author}
\affiliations
KT\\
\emails
\{yg.seo, yoonseok.heo, js.koh, dschang\}@kt.com \\
}

\begin{document}

\maketitle
\begin{abstract}

    In multi-session dialog system, it is essential to continuously update the memory as the session progresses. Simply accumulating memory can make it difficult to focus on the content of the conversation for inference due to the limited input sentence size. Therefore, efficient and accurate conversation model that is capable of managing memory to reflect the conversation history continuously is necessary. This paper presents a conversation model that efficiently manages memory as sessions progress and incorporates this into the model to reflect the conversation history accurately with 3 methodologies: SFT, DPO and DPO with SFT model. Our model using DPO algorithm shows an improvement about 0.0591 of BERTScore in memory accuracy, and the rate of responses reflecting the memory increased as well. Also, response generation performance enhanced about 4.292 in fluency, 3.935 in coherence, and 2.896 in consistency. This paper describes a training method that yields better performance than models with more than twice the parameter size, even when the model size is smaller. Thus, our model demonstrates efficiency not only in terms of accuracy but also in resource utilization.
    
\end{abstract}
\section{Introduction}
The utilization of large-scale language models (LLMs)\cite{gpt4} accompanied by external knowledge has garnered significant attention, as it can overcome the limitations of reusability inherent in lower-level LLMs, which are often associated with high costs. In practice, LLMs embody a vast amount of knowledge across a wide array of topics, unparalleled by past standards. Leveraging their significantly improved contextual understanding, LLMs can maintain consistent and extended conversations with users, addressing various user needs through a single integrated model. However, it is known that LLMs are notably deficient in addressing issues related to knowledge they have not encountered during training or new information generated post-training. This shortfall underscores the fundamental necessity of integrating neuro-symbolic methodologies that leverage up-to-date information sources, such as knowledge graphs or structured databases, during the inference stage of LLMs\cite{ijcai2022p567,ijcai2022p597}.

This paper focuses on the necessity of external knowledge in real industrial environments where small Large Language Models (sLLMs)\cite{phi3_paper} are utilized as conversational agents. Contrary to the scaling law of language models\cite{kaplan2020scaling}, the recently highlighted on-device AI research aims to provide reliable services to users using small-scale language models in constrained environments. Additionally, conversational agents must offer personalized response services to individual users. As this cannot be achieved from the training phase alone, language models must be capable of utilizing individual user information as external knowledge. This user-specific information includes not only meta-data like user preferences but also inferable information from the history of past interactions with the user.

This issue is designed for multi-session dialogue problems, and extensive research has been conducted in this area. Multi-session dialogue assumes situations where conversations occur continuously at regular intervals. A dialogue session is defined as a multi-turn interaction between the user and the system from the initiation to the termination of a conversation. Each dialogue session can range from a few minutes to several months apart. Therefore, the most crucial problem in multi-session dialogue is ensuring that the system can accurately reference information from past dialogue sessions during current interactions.

\begin{figure*}
  \centering
  \includegraphics[width=\textwidth]{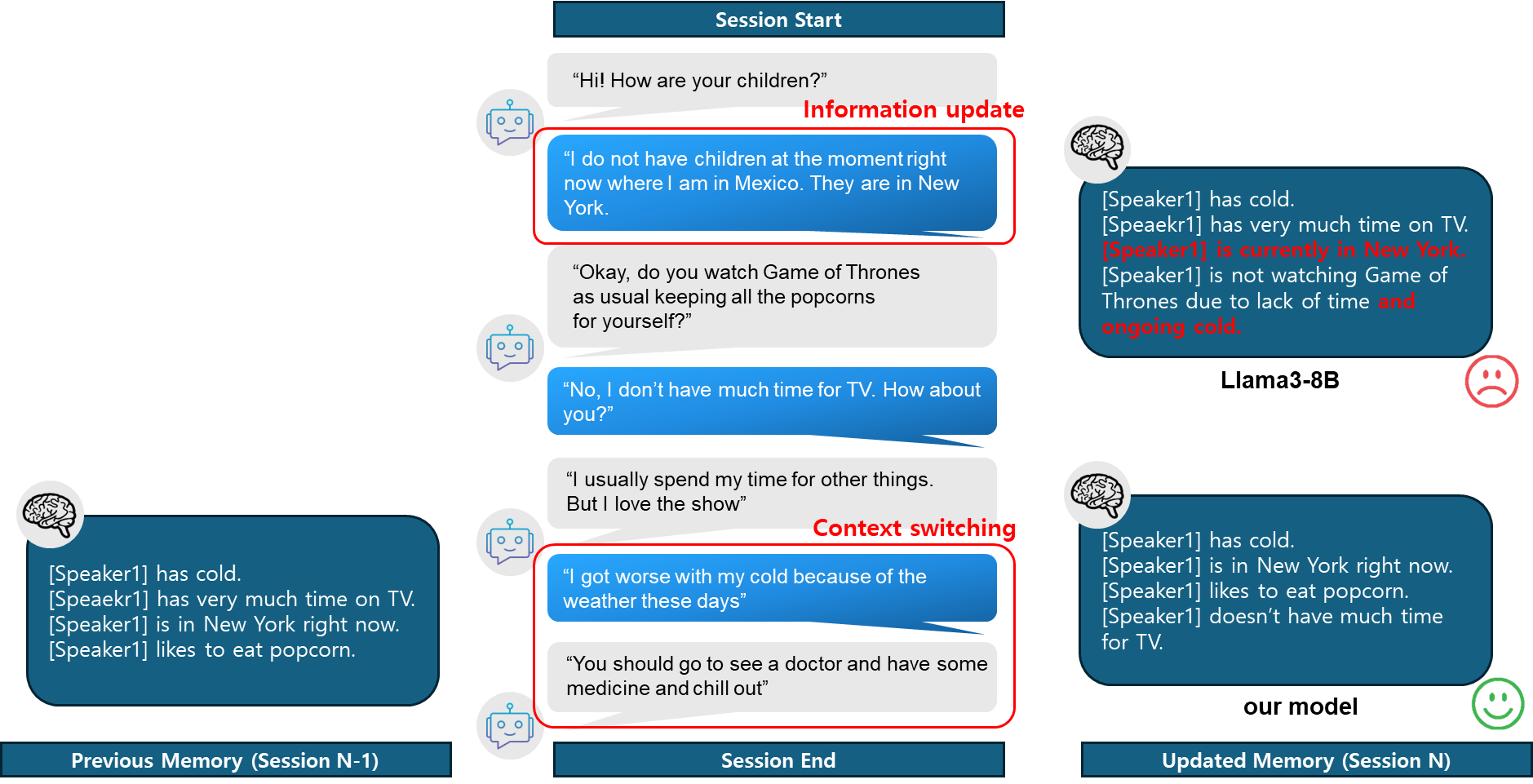}
  \caption{Example of problem in common LLMs}
  \label{fig:intro_figure}
\end{figure*}

Recent research has been directed towards continuously summarizing and updating the history of previous dialogue sessions in the form of sentences, storing and updating them in external memory\cite{wang2024recursively}. As multiple dialogue sessions progress, the information in external memory requires updates. Information in memory may be deleted, updated with new content, or new information may be added over time. Figure \ref{fig:intro_figure} illustrates an example of the memory update process before and after a dialogue session. After the previous session (session N-1), the memory contains information such as speaker1 spending a lot of time watching TV and currently being in New York. However, after the current session concludes, these entries must be updated accordingly.

\cite{wang2024recursively} proposes a methodology that automatically generates the updated memory content itself based on summaries, instead of predicting the operations. This is a straightforward approach that involves tuning GPT-2 models into a summarization problem without any special model modifications, making it more accessible than existing methodologies. However, this research does not exactly match the scenario assumed in this paper, which utilizes sLLM, as the conversation model itself is based on ChatGPT. Therefore, in this paper, we utilized the smaller-scale LLAMA3 model of 8B size, as shown in Figure \ref{fig:intro_figure}. However, as shown in Figure \ref{fig:intro_figure}, two critical issues are inherent such as incorrect memory update from hallucination and misunderstanding of context switching. 

To address these issues, we propose a memory-augmented framework using small language models for consistent multi-session dialogues. We adopt Direct Preference Optimization (DPO)\cite{DPO} to extract and update accurate memory information that reflects causality. Negative samples for DPO training were generated using ChatGPT, constructing prompts to create data that effectively captures causal information in the context. Our proposed framework demonstrates the ability to generate concise summaries and update previous session summaries effectively. Especially our approach is based on a smaller model and memory management, enabling the construction of an efficient and responsive dialogue model.

Our contributions are threefold:
\begin{itemize}
    \item We present a novel pipeline for maintain the memory through multi- session dialog using sLLM. This pipeline leads to figure out the memory and leverages the accuracy of it even though the context switches.
    \item We release automatically augmented DPO negative dataset and prompt for generating such of multi-session dialog memory which can lead to high accuracy.
    \item We demonstrate the effectiveness of DPO algorithm for memory in enhancing the dialog generation for multi-session dialog.
\end{itemize}



\section{Related Work}

\subsection{Memory summarization}
To conduct effective multi-turn or multi-session conversations that accurately reflect the user’s history, it is crucial not only to extract relevant historical information but also to store it efficiently. However, as conversations lengthen and the number of sentences to manage increases, there is a need for efficient processing methods. Previous research\cite{conversation_chronicles} has stored user utterances in a summary format, updating the summary text as sessions progress. However, this method can make memory management challenging due to the increasing length of the summaries.

Other approaches have used hash tables\cite{liu2023thinkinmemory} or transformed summarization into formats like triplets\cite{kang2022knowledgeconsistent} for storage and updates. These transformations, however, can result in the loss of information regarding the relationships between contexts, leading to decreased accuracy. Another study \cite{wang2023augmenting} extracts key-value pairs directly from layers, eliminating the need for separate storage or transformation. However, generating responses that reflect these key-value pairs heavily depends on the model, requiring a pre-trained model from the initial training phase.

In this paper, we propose a model that optimizes resource usage by managing memory through summarization without additional transformations, using only Direct Preference Optimization (DPO)\cite{DPO}. This approach simplifies memory management while maintaining high accuracy and efficiency.

\subsection{Memory management}
While managing memory is essential, the ability to retrieve and update relevant sentences in the conversation is equally important. Utilizing graph embeddings allows for efficient retrieval through hashing algorithms like LSH, but this approach requires converting data into graphs and having a model capable of embedding these graphs. Managing memory through summarization by structuring it into short chunks in a list format simplifies updates. However, this method necessitates clear criteria for replacing, adding, or deleting sentences, which typically involves training additional models like NLI for effective management.

By employing prompts to let large language models (LLMs) make these decisions mechanically, memory updates can be achieved without additional model training. This paper proposes an enhanced method that applies reinforcement learning to improve the accuracy of memory updates within given dialogues. To address the challenge of reflecting context switching, we automatically generate negative samples and use them with the Direct Preference Optimization (DPO) algorithm to focus on the context.

This approach not only simplifies the process but also ensures that memory updates are precise and contextually relevant, leveraging advanced techniques to overcome the limitations of previous models.

\begin{figure*}[ht]
  \centering
  \includegraphics[width=\textwidth]{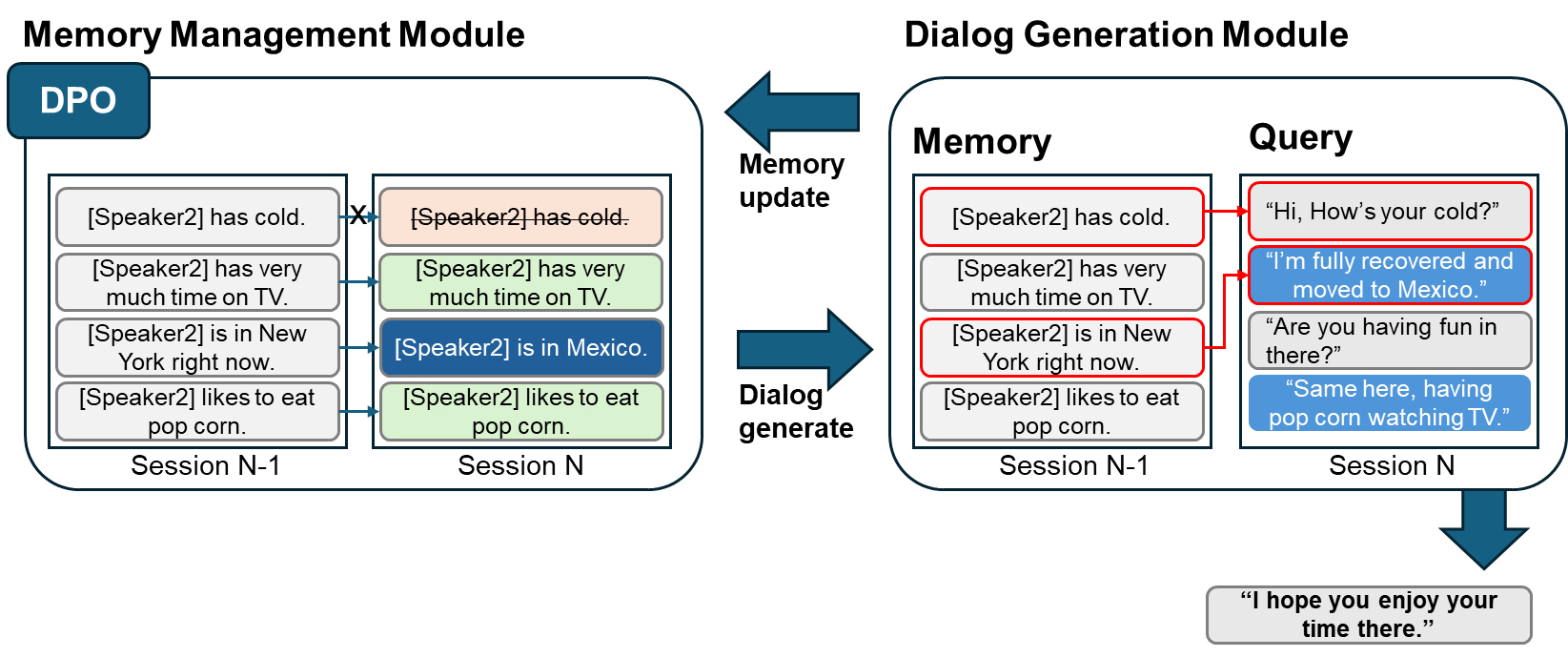}
  \caption{A proposed framework for memory-augmented small language models}
  \label{fig:main_figure}
\end{figure*}

\section{Methodology}

In this section, we propose a memory-augmented framework using small language models for consistent multi-session dialogues. As shown in Figure \ref{fig:main_figure}, it consists of two modules: memory management module and dialog generation module. 

The primary purpose of the memory management module is to store speaker utterance information from previous conversation sessions in a separate memory space for reference in future new conversation sessions. This module updates the memory by referencing the contents of the current conversation session based on the speaker information updated up to the previous conversation session immediately after each conversation session ends. To achieve this, we propose a history management method based on summary using a DPO-tuned SLM. The proposed method allows for managing the key information of speaker utterances in the form of a set of short sentences. This approach has the advantages of improving the accuracy of generated responses, maximizing learning efficiency, and facilitating easy monitoring through the management of speaker information in a concise format.

Next, the goal of the second conversation generation module is to provide the user with more consistent and accurate generated responses by effectively reflecting not only the current session's context but also the information from previous conversation sessions. As shown in Figure \ref{fig:main_figure}, the conversation generation module receives information from the memory updated up to the (N-1)th conversation session as input prompts. This enables the module to appropriately reference conversation history that occurred in the past, thereby addressing past issues arising from temporal differences. In this paper, we employ the phi-3 as the conversation generation module without additional training. Consequently, we observed that it possesses a higher generation accuracy compared to the llama3-8b model, which has over twice the number of parameters. The following subsections provide a detailed description of each module, with particular emphasis on the learning method of the effective memory management module utilizing DPO, which is the distinct aspect of this paper.

\begin{figure*}[ht]
  \centering
  \includegraphics[width=\textwidth]{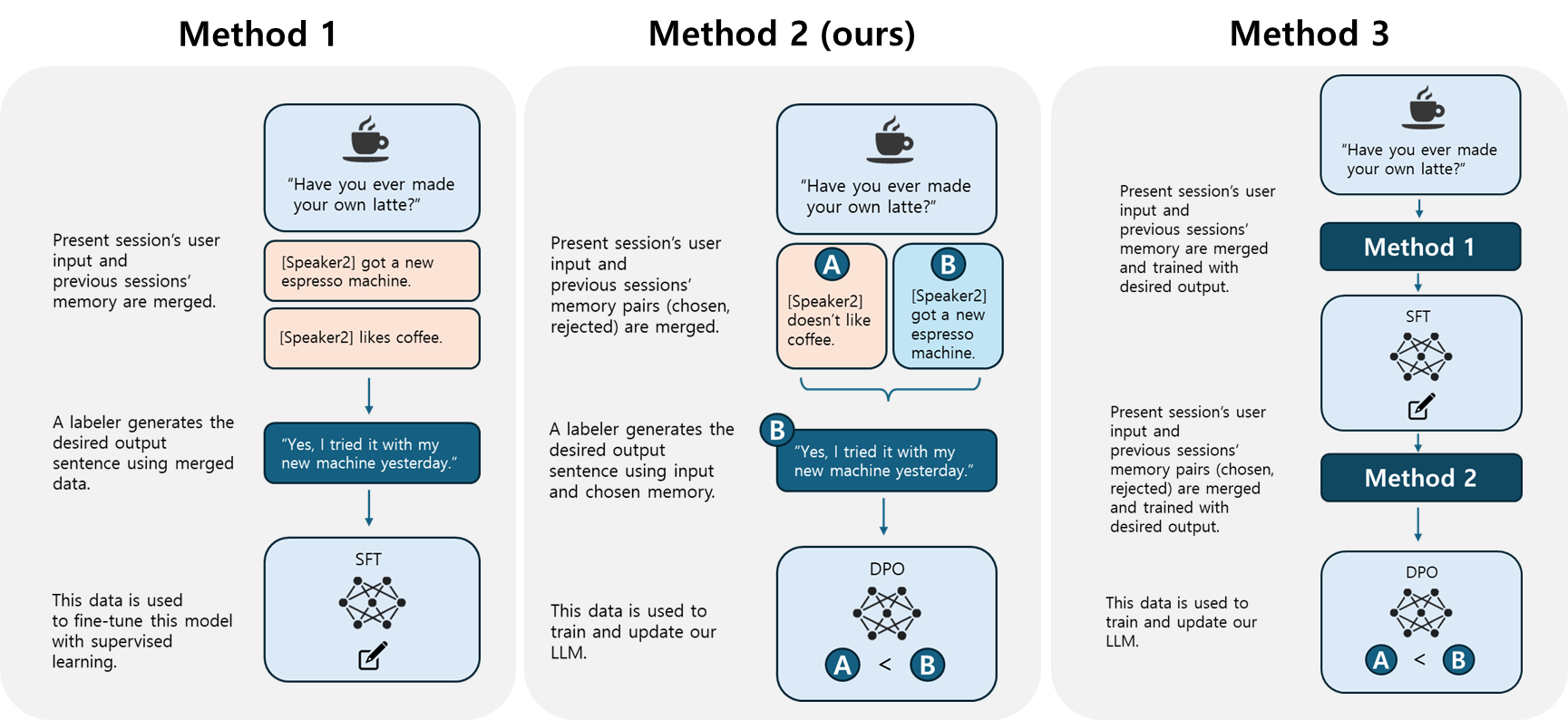}
  \caption{Overview of the methodology of training models.}
  \label{fig:training_scheme}
\end{figure*}

In this paper, we designed 3 methods for memory management as SFT, DPO and DPO with SFT model. The details of each training methods are presented in Figure \ref{fig:training_scheme}. After the memory is summarized with such methods, the response is generated with the opening question of the next dialog session using the updated memory. The example of the memory management and dialog generation pipeline is shown in Figure \ref{fig:main_figure}.

\subsection{Memory Management Module}
In this module, the context of the dialog is summarized separately by speakers and stored in list format as memory to manage memory efficiently at the end of the session's dialog. As the session continues, in the case of multi-session conversations, the stored memory is updated with information necessary for the following conversations. When the memory is updated, changes are easily deleted or replaced as it is organized in a list format. This pipeline allows for more accurate information when generates the response.

\subsubsection{Training Methods}
We introduce a training scheme for the memory management module by only adopting the DPO algorithm, which is denoted as Method2. To make readers better understand, we introduce different training schemes such as SFT and SFT with DPO. 

\indent\textbf{Method1: SFT (Supervised fine-tuning)} \\
Supervised fine-tuning is a machine learning technique where a pre-trained model is further trained on a specific task using labeled data. The pre-trained model, often trained on a large dataset for a related task, serves as a starting point, and then the model's parameters are adjusted (fine-tuned) to better fit the new task-specific data.

In supervised fine-tuning, the model's parameters are updated to minimize a predefined loss function, typically a measure of the difference between the model's predictions and the ground truth labels. This process allows the model to adapt its learned representations to the specific characteristics of the new task while leveraging knowledge gained from the pre-training phase. Supervised fine-tuning is commonly used in transfer learning scenarios, where the pre-trained model's knowledge is transferred to new tasks, leading to improved performance, especially when labeled data for the target task is limited.

\indent\textbf{Method2: DPO (Direct Preference Optimization)} \\
Reinforcement learning is a machine learning paradigm where an agent learns to interact with an environment to achieve a goal by maximizing cumulative rewards. The agent takes actions in the environment, observes the resulting state transitions and received rewards, and learns a policy that maps states to actions to maximize long-term rewards.

Supervised fine-tuning is a form of supervised learning, where the model learns from labeled data with a predefined objective, whereas reinforcement learning is a type of learning where the agent learns to interact with an environment to maximize cumulative rewards through trial and error.

In particular, the method utilized in this paper aims to utilize DPO to learn which preference, either positive or negative sample, should be selected, thereby reducing hallucinations and focusing on learning within the given conversational context. Through this approach, we aim to apply a learning method that can effectively understand the cause and effect of the context using simple pairwise comparisons, requiring fewer resources compared to SFT learning methods that label data extensively. 

\indent\textbf {Method3: SFT and DPO}
Using a model trained with supervised fine-tuning (SFT) to perform direct preference optimization (DPO) would likely yield performance enhancement. The model enhanced through SFT might have already acquired rich knowledge from various datasets. This enhanced model could improve performance when used in DPO tasks where additional training data is limited. It could maintain pre-learned features while enhancing performance in predicting human preferences. Also, DPO relies on human preferences rather than predefined reward functions. The model improved through SFT might be more effective in predicting these human preferences.

Therefore, utilizing this model could result in more accurate and efficient estimation of human preferences. This could facilitate quicker adaptation to human preferences and enable faster decision-making processes.

\subsection{Dialog Generation Module}
Dialog generation module uses the memory generated from Memory management module and the context of the current session to generate responses to the user's dialog. In this paper, we use Phi-3 as the base model. The example of prompt for the model is described in Figure \ref{fig:prompt_example}. If the memory is well-structured, any language model can be used for the base model of generating the conversation in dialog generation module. By combining the memory sentence list and dialog context to generate responses, the module is able to store previous sessions' memory providing information more focused on the user.

\begin{figure}[ht!]
    \centering
    \begin{boxB}
    Answer as [Speaker2] after the dialog using the memory of the previous session. \textbackslash n\textbackslash n Previous memory: \{memory\} \textbackslash n\textbackslash n Dialog context: \{context\}
    \end{boxB}
    \caption{Example prompt of Dialog Generation Module}
    \label{fig:prompt_example}
\end{figure}

\section{Experiment}
 In this section, we present the experimental results evaluating the summarization and summarization update processes for memory management. We describe examples of the prompts used for each evaluation or data generation and their respective outcomes. The evaluations were conducted using both auto-evaluation with ChatGPT and human evaluation. Utilizing these methods, we compare and discuss the performance of our model relative to other models.


\subsection{Experiment Details}

\subsubsection{Dataset}
 To store dialogue in memory in a suitable summarization format, it is essential to extract speaker information that can be utilized in subsequent sessions. Therefore, we utilize Multi-Session Chat(MSC,Xu et al.,2022a) dataset which has summary information of each speaker as persona, and aggregated summary information after the following sessions. To facilitate continuous updates of the summarized information, we structured the summaries into short sentences of 15 words or fewer, making them easier to manage than longer context sentences. To effectively train this approach, we designed prompts to create a dataset for Supervised Fine-Tuning (SFT).

 Table \ref{tab:dataset_details} presents the details of each train data and test data that are used for the methods.
\begin{table}[h]
\centering
\caption{Dataset details of train and test data}
\label{tab:dataset_details}
\begin{tabularx}{\columnwidth}{Xr}
    \toprule
    \textbf{Dataset} & \textbf{Number of Cases} \\
    \midrule
    Train dataset (SFT)          & 2,000              \\
    Train dataset (DPO)         & 2,000         \\
    Test dataset             & 1,000              \\
    \bottomrule
\end{tabularx}
\end{table}

\subsubsection{DPO negative sample}
 To apply Direct Preference Optimization (DPO), it is necessary to have pairs of negative and positive samples consisting of rejected and selected samples. The reason for applying DPO in this paper is to accurately reflect the causal facts in the memory. Previous research has used a method of generating negative samples by altering some entities in given sentences to develop benchmarks for evaluating factuality [10]. Therefore, we used prompts to select sentences that accurately reflect cause and effect, generating suitable negative samples for reinforcement learning.


\begin{figure}[ht!]
    \centering
    \begin{boxB}
    Write diverse alternative sentence that contradicts the original case or result by modifying an entity. Sentence: \{sentence\}
    \end{boxB}
    \caption{Example of generating DPO negative sample prompt}
    \label{fig:DPO_prompt}
\end{figure}

\subsubsection{Evaluation Metric}

 In order to quantitatively compare the prediction sentences and target sentences pairwise, we assessed the quality of the summary using the following metrics. The prediction sentences are those generated by the model, while the target sentences refer to the list of correct sentences that need to be stored in memory from a given session. Predicted summary sentences that exceed a certain threshold are considered Similar sentences. Thus, the quality of the summary can be evaluated based on how many of the target sentences are matched by the prediction sentences above a specific threshold. For evaluating the quality of the summary, we employed cosine similarity and BERTScore, which are commonly used to assess the similarity between two sentences. The metric is described in Figure \ref{fig:metric_example}.

 Given the target dataset $T= \left ( \! t_{1}, t_{2}, t_{3} \right )$ where each of the item represents the answer summary of the memory, our goal is to evaluate the pair between the target and the given prediction dataset $P= \left ( \! p_{1}, p_{2}, p_{3} \right )$. Here, cosine similarity and BERTScore are used to calculate the similarity between the each pair of target and prediction. When the similarity value is more than 0.7 and 0.95 respectively, the sentences are expected to be similar and represented as  $S= \left ( \! s_{1}, s_{2} \right )$. Therefore, the accuracy of the summarization refers to the ratio of the number of target sentences to the number of similar sentences that are expected to be the answers.

\begin{figure}[ht!]
    \centering
    \begin{boxB}
    $T= \left ( \! t_{1}, t_{2}, t_{3} \right )$, $P= \left ( \! p_{1}, p_{2}, p_{3} \right )$, $S= \left ( \! s_{1}, s_{2} \right )$ \newline
    $S_{1}=cosine\_similarity(t_{n}, p_{m}) \geq  0.7$,\newline
    $S_{2}=BERTScore(t_{n}, p_{m}) \geq  0.95$ \newline
    $Accuracy=n(S_{1})/n(T)$  or $n(S_{2})/n(T)$
    \end{boxB}
    \caption{Detail of metric}
    \label{fig:metric_example}
\end{figure}

\subsubsection{Automatic Evaluation}
 We conduct an automatic evaluation using GPT-4o, the latest model introduced by OpenAI. The evaluation prompt is described in Figure \ref{fig:geval_example}.

\begin{figure}[ht!]
  \centering
  \begin{boxB}
    You're an impartial judge. You will be shown a Conversation Context, Personality of Speakers and Assistant Response. \newline
    \#Fluency: Please evaluate whether the Assistant's response is natural, fluent and similar to human communication, avoiding excessive repetition and ensuring a diverse range of output. \newline
    \#Consistency: Please evaluate whether the Assistant's response is consistent with the information of persona list. Any deviation from the expected personality may indicate a lack of coherence. \newline
    \#Coherency: Please evaluate whether the Assistant's response maintains a coherent and logical flow of conversation based on the evolving context. A response with good context coherence can understand and respond appropriately to changes in conversation topics, providing smooth and sensible interactions. \newline \newline
    Conversation Context: \{dialog\} \newline
    Personality: \{persona\} \newline
    Assistant Response: \{response\} \newline
    Rate the Assistant Response on an integer score of 1 (very bad) to 100 (very good).
  \end{boxB}
  \caption{Example of evaluation prompt}
  \label{fig:geval_example}
\end{figure}

\noindent We follow the G-Eval methodology \cite{liu2023geval} for conducting automatic evaluations.

\subsubsection{Baselines}
 To demonstrate that the model utilizing sLLM outperforms the LLM model and generates more accurate responses, we conducted experiments using the following models:

\begin{itemize}
\item 
\textbf{Phi.} It is an LLM with a very small size of 3.8B parameters. This model was used to summarize previous sessions and store them in memory for updates. The purpose was to show that even with a smaller parameter size, an sLLM can sufficiently reflect the information from previous sessions.
\item 
\textbf{Llama-3.} We use this model as the base for generating dialog responses. To ensure effective integration with memory information, we experimented with an 8B parameter model.
\end{itemize}

\subsection{Result}

 First, to assess the dialog generation performance when utilizing memory updates, we conducted a multi-session configuration. In this setup, the summary generated from the first session was used as memory, and the ability of different models to respond to the user's opening question in the next session was evaluated. This allows us to determine whether the models can generate responses that reflect the memory from previous sessions. The results comparing the use and non-use of memory updates are shown in the following table. Specifically, it demonstrates that incorporating the memory from previous sessions leads to more accurate results in subsequent sessions.

\subsubsection{Evaluation of Memory summarization}
 To compare the summary performance of each model, we trained the models using their respective methods and evaluated the output in list format against the actual summary output. We utilized commonly used metrics such as BLEU, ROUGE, F-1, and BERTScore to compare performance as shown in Table \ref{tab:comparsion_metric_1}. Instead of merely counting matching words, BERTScore measures semantic similarity between sentences using BERT embeddings. This ensures meaningful comparisons that take context into account. 

 It shows much better performance about 0.1267 higher performance after trained with DPO algorithm with phi-3 model itself and enhanced about 0.0591 in BERTScore and more accurate about 0.0489 than Llama3-8B. Upon the results, it was observed that the proposed model trained with DPO exhibited the best performance. Particularly, compared to model\_1 trained directly for summarization through SFT, the performance of our model is approximately 0.0399 higher. Furthermore, considering the significant difference in parameter size, with the Llama model having over twice the parameters, the observed increase in performance suggests that the DPO-based learning method is highly efficient.

\begin{table*}[]
\centering
\caption{Comparison of memory summarization performance}
\label{tab:comparsion_metric_1}
\begin{tabular}{llllll}
\hline
\textbf{} & \textbf{model\_1} & \textbf{\begin{tabular}[c]{@{}l@{}}model\_2\\ (ours)\end{tabular}} & \textbf{model\_3}  & \textbf{Llama3-8B} & \textbf{phi-3} \\ \hline
BLEU-1    & 0.1048            & \textbf{0.1266}           & 0.0972                                                             & 0.0785    & 0.0892         \\ 
BLEU-4    & 0.0056            & 0.0051            & \textbf{0.0052}                                                    & 0.0048             & 0.0035         \\
ROUGE-L   & 0.1086            & \textbf{0.1208}            & 0.1110                                                             & 0.0971    & 0.0962         \\
F-1       & 0.1120            & \textbf{0.1280}            & 0.0986                                                             & 0.1211    & 0.0898         \\
BScore    & 0.9012            & \textbf{0.9411}            & 0.8922                                                    & 0.8735             & 0.8820         \\ 
human evaluation & 0.4250 & \textbf{0.6750} &0.5730 & 0.5250 & 0.4350 \\ \hline
          &                   &                   &                                                                    &                    &               
\end{tabular}
\end{table*}

\subsubsection{Evaluation of Memory update}

 To figure out performance of how the memory is managed with the each of methodologies, the evaluation of memory update performance is needed. The each model predicts the aggregated or updated memory with the previous session memory and the present session's dialog.
As demonstrated by the results in Table \ref{tab:comparsion_metric_2}, the summarization performance of the Phi-3 model, with roughly half the parameter size, is comparable to or even better than that of the Llama-3 8B model. However, using a simple reference and prediction comparison based on n-grams to evaluate summarization may not be appropriate. Particularly, the summarization predictions generated by our model are output in list format rather than continuous text, making it unsuitable to compare using newline characters and similar tokens. Therefore, our model's superior performance is highlighted by achieving higher scores on BScore, a metric that reflects word similarity more comprehensively than n-grams.


\begin{table*}[h]
\centering
\caption{Comparison of memory update performance}
\label{tab:comparsion_metric_2}
\begin{tabular}{llllll}
\hline
\textbf{} & \textbf{model\_1} & \textbf{\begin{tabular}[c]{@{}l@{}}model\_2\\ (ours)\end{tabular}} & \textbf{model\_3}  & \textbf{Llama3-8B} & \textbf{phi-3} \\ \hline

BLEU-1    & 0.0754            & \textbf{0.1057}            & 0.0729                                                             & 0.0847    & 0.0702         \\ 
BLEU-4    & 0.0048            & \textbf{0.0052}            & 0.0041                                                    & 0.0045             & 0.0037         \\
ROUGE-L   & 0.0093            & \textbf{0.1118}            & 0.0946                                                             & 0.1021    & 0.0960         \\
F-1       & 0.0976            & 0.1066            & 0.0946                                                             & \textbf{0.1250}    & 0.0977         \\
BScore    & 0.8618            & \textbf{0.8662}            & 0.8610                                                    & 0.8538             & 0.8609         \\ 
human evaluation & 0.3750 & \textbf{0.6250} & 0.3750 & 0.5000 & 0.2500 \\ \hline
          &                   &                   &                                                                    &                    &               
\end{tabular}
\end{table*}

 Using quantitative metrics has limitations in determining whether the memory is accurately reflected in the generated sentences. Therefore, we sampled the evaluation dataset for human evaluation as shown in Table \ref{tab:comparsion_metric_2}. In many cases, the accuracy of summarization decreased due to swapped information between speaker1 and speaker2 or confusion in the context of the actual dialogue. However, the model trained with DPO demonstrated a higher accuracy in extracting information within the context.

 The results of applying the pairwise comparison method proposed in this paper, which accounts for list-formatted prediction outputs, are presented in the following table. In Table \ref{tab:comparsion_metric_2}, it shows clear comparison between the proposed models and others. The predictive performance of the proposed model is illustrated with examples. Overall, the proposed model achieved the highest summary reflection ratio of 0.8722 indicating that the update of the memory with the proposed algorithm works well with summarization and update as well.

\subsubsection{Evaluation of Dialog generation}
 After the model updates the previous memory according to the present session, our final goal for the pipeline is to generate the response of the session's dialog using it. Therefore, the output response of the user's question in the preis compared after the memory update. When the model generates the output response, updated memory is combined with the user's question. We check the accuracy of the model with 3 different aspects; Fluency, Coherence, and Consistent. The automatic evaluation is used for each of the aspects and it shows that model\_2 has the most fluent, coherent and consistent dialog generation ability at the same time. The prompt that used for the evaluation is shown in Figure \ref{fig:geval_example}.
 Notably, the results show that DPO training with negative samples yields better performance about 3.931, 4.496 and 2.875 more than average compared to training with SFT or SFT with DPO afterwards as shown in Table \ref{tab:comparsion_fcc}. Similar to the result of the memory update, the accurate memory naturally leads to the high performance of dialog generation. Especially, it is interesting that not only consistency increases, but also even the fluency of the model response is enhanced with DPO algorithm. 


\begin{table}[H]
\centering
\caption{Comparison of Dialog generation performance}
\label{tab:comparsion_fcc}
\begin{tabular}{lcccc}
\toprule
Method & Flu. & Coh. & Cons. & Overall \\
\midrule
Model\_1 & 84.948 & 80.031 & 83.021 & 82.667 \\
Model\_2 (ours) & \textbf{92.750} & \textbf{89.427} & \textbf{90.406} & \textbf{90.083} \\
Model\_3 & 89.427 & 84.521 & 87.229 & 87.392 \\
Llama3-8b & 91.552 & 80.656 & 89.489 & 87.965 \\
Phi-3 & 88.458 & 85.479 & 87.510 & 87.149 \\
\bottomrule
\end{tabular}
\end{table}

\section{Conclusion} 
This paper describes a training method that yields better performance than models with more than twice the parameter size, even when the model size is smaller, by experimenting with three methods. The methods include SFT (Supervised Fine-Tuning), DPO (Direct Preference Optimization), and DPO training using an SFT model. The analysis of results demonstrates that DPO, in particular, allows for the most efficient memory update and enhances dialogue generation performance. By learning pair sentences that include negative sets to extract cause and effect information from the context, DPO outperforms larger models. Furthermore, it was observed that generating responses based on accurate information leads to better answers, even as the session length increases.

\bibliographystyle{named}
\bibliography{ijcai24}

\end{document}